\documentclass[journal]{IEEEtran}
\usepackage{multirow}
\usepackage{amssymb,amsmath}
\usepackage{rotating}
\usepackage[ruled,vlined]{algorithm2e}
\usepackage{algorithmic}
\usepackage{bm}
\usepackage{graphicx}
\usepackage{subfigure}
\usepackage{color}
\usepackage{amsthm}
\usepackage{amsbsy}

\usepackage{tabularx}
\usepackage{multirow}

\ifCLASSINFOpdf
\else
\fi
\begin{document}
\title{Fine-Grained Age Estimation in the Wild with Attention LSTM Networks}

\author{Ke~Zhang,~\IEEEmembership{Member,~IEEE,}
        Na~Liu,
        Xingfang~Yuan,~\IEEEmembership{Student Member,~IEEE,}
        Xinyao~Guo,
        Ce~Gao,
        Zhenbing~Zhao,~\IEEEmembership{Member,~IEEE,
        and~Zhanyu~Ma,~\IEEEmembership{Senior Member,~IEEE}
        }% <-this % stops a space
\thanks{This work is supported by the National Natural Science Foundation of China (Grant Nos. 61871182, 61773071, 61302163 and 61302105), Hebei Province Natural Science Foundation (Grant Nos.F2015502062 and F2016502062) and the Fundamental Research Funds for the Central Universities (Grant No.2018MS094, 2018MS095 and 2018XKJC02). The authors gratefully acknowledge the support of NVIDIA Corporation with the kind donation of the GPU used for this research.}% <-this % stops a space
\thanks{K. Zhang is with the Department
of Electronic and Communication Engineering, North China Electric Power University, Baoding,
Hebei, 071000 China, e-mail: zhangkeit@ncepu.edu.cn.}% <-this % stops a space
\thanks{N. Liu is with the Department
	of Electronic and Communication Engineering, North China Electric Power University, Baoding,
Hebei, 071000 China, e-mail: liuna6882@126.com.}% <-this % stops a space
% <-this % stops a space
\thanks{X. Yuan is with the Department
	of Electrical and Computer Engineering, University of Missouri, Columbia,
	MO, 65211 USA, e-mail: xyuan@mail.missouri.edu.}% <-this % stops a space
\thanks{X. Guo is with the Department
	of Electronic and Communication Engineering, North China Electric Power University, Baoding,
	Hebei, 071000 China, e-mail: ggzxyw@gmail.com.}% <-this % stops a space
\thanks{C. Gao is with the Department
  of Electronic and Communication Engineering, North China Electric Power University, Baoding,
  Hebei, 071000 China, e-mail: 940770901@qq.com.}% <-this % stops a space
\thanks{Z. Zhao is with the Department
  of Electronic and Communication Engineering, North China Electric Power University, Baoding,
  Hebei, 071000 China, e-mail: zhaozhenbing@ncepu.edu.cn.}
\thanks{Z. Ma (Corresponding author) is with the Pattern Recognition and Intelligent System Laboratory, Beijing University of Posts and Telecommunications, Beijing,
  100876 China, e-mail:mazhanyu@bupt.edu.cn.}}% <-this % stops a space
\maketitle

\begin{abstract}
Age estimation from a single face image has been an essential task in the field of human-computer interaction and computer vision, which has a wide range of practical application values. Accuracy of age estimation of face images in the wild is relatively low for existing methods, because they only take into account the global features, while neglecting the fine-grained features of age-sensitive areas. We propose a novel method based on our attention long short-term memory (AL) network for fine-grained age estimation in the wild, inspired by the fine-grained categories and the visual attention mechanism. This method combines the residual networks (ResNets) or the residual network of residual network (RoR) models with LSTM units to construct AL-ResNets or AL-RoR networks to extract local features of age-sensitive regions, which effectively improves the age estimation accuracy. First, a ResNets or a RoR model pretrained on ImageNet dataset is selected as the basic model, which is then fine-tuned on the IMDB-WIKI-101 dataset for age estimation. Then, we fine-tune the ResNets or the RoR on the target age datasets to extract the global features of face images. To extract the local features of age-sensitive regions, the LSTM unit is then presented to obtain the coordinates of the age-sensitive region automatically. Finally, the age group classification is conducted directly on the Adience dataset, and age-regression experiments are performed by the Deep EXpectation algorithm (DEX) on MORPH Album 2, FG-NET and 15/16LAP datasets. By combining the global and the local features, we obtain our final prediction results. Experimental results illustrate the effectiveness and robustness of the proposed AL-ResNets or AL-RoR for age estimation in the wild, where it achieves better state-of-the-art performance than all other convolutional neural network (CNN) methods on the Adience, MORPH Album 2, FG-NET and 15/16LAP datasets.
\end{abstract}


% \fi
\IEEEpeerreviewmaketitle

\section{Introduction}
\begin{figure}[t]
\centering
\includegraphics[width=1.0\linewidth]{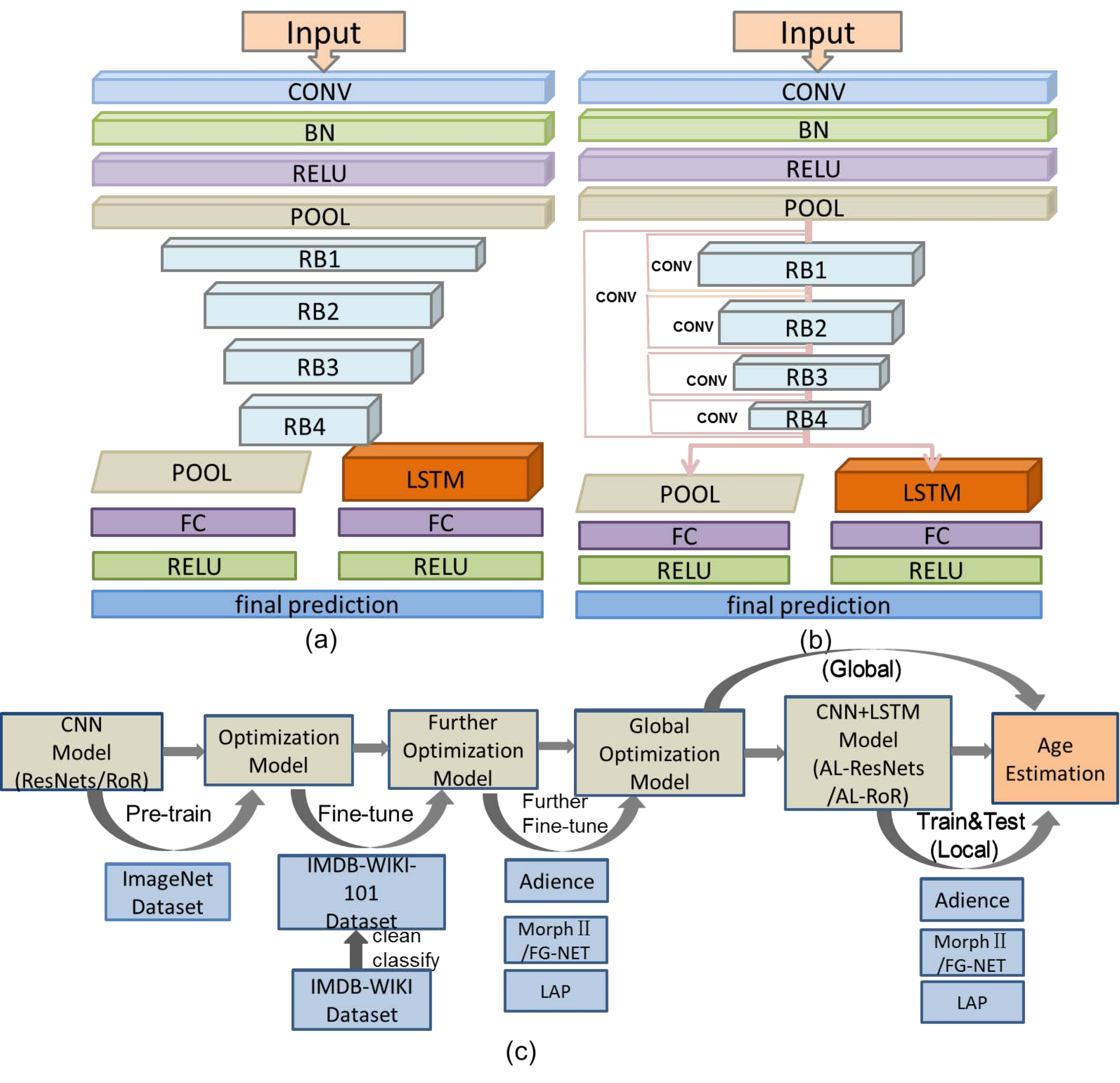}
\caption{Fig.~\ref{fig:basicframework}(a, b) are AL-ResNets and AL-RoR architectures, where one LSTM unit is integrated into the original ResNets and RoR. Fig.~\ref{fig:basicframework}(c) is the pipeline of our framework for fine-grained age estimation. The ResNets or RoR model is pretrained on ImageNet and IMDB-WIKI-101 datasets to obtain the optimized model first, and then fine-tuned on the target age datasets to obtain the global features. Finally, the local features of the age-sensitive regions are extracted by AL-ResNets or AL-RoR network and combined with the global features to get our final prediction results.}
\label{fig:basicframework}
\end{figure}
\IEEEPARstart{A} variety of face attributes from face images in the wild are known to be very useful in characterizing the individual characteristics. Age is one of the significant properties, which is regarded as an inherent attribute and a crucial biological characteristic, which plays a fundamental role in human social interaction. Therefore, automatic age estimation from face images is an important research content in the field of artificial intelligence. Age estimation is an available technique for many applications, which are depends on accurate age information, such as medical diagnosis (premature facial aging due to various causes), age-based human-computer interaction system, advanced video surveillance, demographic information collection, soft biometrics, etc.
\par
Most previous studies addressed the age prediction problem using hand-designed features with statistical models~\cite{ref-a}~\cite{ref-b}. Although many traditional statistical models~\cite{ref-c} have been proposed, hand-designed features behave unsatisfactorily on benchmarks of unconstrained images. Later research approached age estimation from face images in convolutional neural network (CNN) manner, automatically extracting feature representations for input images. There are two reasons  why automatic age estimation is regarded as a very challenging task. First, discriminative feature extraction for age estimation is easily affected by large variations in facial gestures, lighting, makeup, background, noise, etc.~\cite{ref-1}. Second, the significant similarity and subtle inter-class differences in face images with adjacent ages are hardly handled. Therefore, distinguishing age with only global features of face may not achieve better results, while searching for age-sensitive regions (wrinkles, hair, liver spots, etc.) can provide more distinctive features for age estimation. Inspired by fine-grained image categories and attention conceptions~\cite{ref-d}, we argue that combining the local features of age-sensitive regions with global features may be helpful to age estimation. In contrast to traditional fine-grained category datasets, where manual annotation of a specific part is the dominant strategy for the target, a wide array of fine-grained classification methods exist~\cite{ref-2}~\cite{ref-3}~\cite{ref-4}~\cite{ref-5} and are still deployed with additional supervisory information which increases computational complexity. Currently, none of the several public large age datasets mark certain image parts, which severely limits the development of fine-grained age estimation. Therefore, there is an urgent demand on how to automatically obtain position information of age-sensitive regions.

\par
To solve such problems, we propose a method of fine-grained age estimation based on our attention LSTM (AL) network to improve the abilities of feature extraction. As shown in Fig.~\ref{fig:basicframework}(a, b), the long short-term memory (LSTM) unit is seamlessly inserted between the residual group and the fully connected layer of the residual network (ResNets) or the residual network of residual network (RoR) to form an AL-ResNets or AL-RoR network. The network is designed to effectively combine the ResNets or RoR with LSTM unit to generate feature representations on the critical age-sensitive regions. In the context of fine-grained age estimation, extracting features can be regarded as a two-level process, where one is image-level and the other one is part-level.
\par
Fig.~\ref{fig:basicframework}(c) shows the pipeline of our framework. First, to improve the performance and alleviate the over-fitting problem on small-scale datasets, we train ResNets or RoR model on ImageNet~\cite{ref-00}; then we fine-tune it on the IMDB-WIKI-101 dataset~\cite{ref-22}. Second, we use the model to further fine-tune it on target age datasets to extract the global features of the images. Third, on the premise of image-level features extraction, an AL-ResNets or AL-RoR network based on ResNets or RoR is constructed, where the local features of the age-sensitive regions on target images are extracted; then the final prediction results are obtained by combining the predictions of the global and local features together. Finally, through abundant experiments on several popular age datasets, our models achieved new state-of-the-art results on Adience~\cite{ref-1}, MORPH Album 2~\cite{ref-31}, FG-NET~\cite{ref-20} and 15/16LAP datasets~\cite{ref-46}~\cite{ref-56}.
\par
Our main contribution is threefold:
\par
1. Our method employs fine-grained categories and visual attention mechanism in the age estimation field for the first time. The proposed Attention LSTM network extracts local age-sensitive regions and more distinctive features automatically, which can effectively improve the accuracy of age estimation.
\par
2. The proposed Attention LSTM network has few extra parameters compared with the original CNN models and is easily trained, so our method is not only effective but also practical for age estimation.
\par
3. Through massive experiments, we analyzed the effects of different training strategies, network structures (ResNets and RoR) and network depths for age estimation, then developed reasonable strategies and achieved the new state-of-the-art results on different datasets. Moreover, the results of experiments based on ResNets and RoR verified the robustness of our method with different network structures.
\par
The rest of the paper is organized as follows. Section II briefly reviews related work for age estimation methods. Section III describes the proposed AL-ResNets or AL-RoR age estimation method. Section IV presents experimental results and analysis leading to conclusions in Section V.

\section{Related Work}
\par
Automatic face analysis is a research topic that is currently receiving much attention from the computer vision and pattern recognition communities. Age has been investigated as a soft biometric~\cite{ref-a} and facial attribute~\cite{ref-62}, and age estimation has historically been one of the most challenging problems within the field of facial analysis. Age estimation used to extract facial features manually in the past, but now CNN methods~\cite{ref-22} are preferred due to achievements in training CNN directly on age datasets. Face age datasets are divided into biological age datasets and apparent age datasets, and diverse age datasets can apply different age estimation methods. Much research has been devoted to age estimation from a face image under the more familiar biological age estimation. Adience, MORPH Album 2 and FG-NET are prevalent benchmarks for biological age estimation; their image labels are marked by the age group or actual age; thus the predicted output is the biological age of a person. In contrast, apparent age estimation research is still in its infancy. Only one publicly available dataset is used in the context of apparent age estimation: Chalearn LAP including 15LAP and 16LAP. What distinguishes it from other datasets is that the age of each image in this dataset is labeled by the average of annotators' subjective opinions, so apparent age is the visual age based on the perspective of human.
\subsection{Biological Age Estimation}

%\IEEEpubidadjcol
In the past 20 years, biological age estimation from face image has benefited tremendously from the evolutionary development in facial analysis. Based on hand-designed features, regression and classification methods were used to predict the age of face images. AGing pattErn Subspace(AGES)~\cite{ref-23} was constructed to model the aging pattern, which was implemented for automatic age estimation. Subsequently, Chang et al.~\cite{ref-24} proposed an ordinal hyperplane ranking algorithm called ordinal hyperplanes ranker (OHRank) for estimating human age via facial images. Wang et al.~\cite{ref-25} proposed a new framework for age feature extraction based on a manifold learning algorithm and the deep learned aging pattern (DLA), which greatly improved the age estimation performance. Chen et al.~\cite{ref-26} proposed a cumulative attribute concept based on support vector regression (SVR) for learning a regression model, and it gained a notable advantage based on its accuracy for age estimation.  Guo et al.~\cite{ref-28} proposed a kernel canonical correlation analysis (KCCA) method, which could derive an extremely low dimensionality in estimating age, but the amount of kernel calculation was tremendous. All of these methods had the same scope of applicability, which was only proven effective on constrained benchmarks but could not achieve acceptable results on benchmarks for images captured in the wild.
\par
Recent research on CNN showed that CNN models~\cite{ref-7}~\cite{ref-9}~\cite{ref-11}~\cite{ref-12}~\cite{ref-13}~\cite{ref-14}~\cite{ref-15}~\cite{ref-61} could learn compact and discriminative feature representations when the size of training data is sufficiently large, so an increasing number of researchers have started to use CNN for age estimation. Levi et al.~\cite{ref-16} applied DCNN for the first time to age classification on an unconstrained Adience benchmark. Yi et al.~\cite{ref-29} proposed a multi-scale convolution neural network based on the traditional face analysis method. Hou et al.~\cite{ref-17} proposed a VGG-16 model with smooth adaptive activation function (SAAF) to predict age group on the Adience benchmark. Then they used the exact squared Earth Movers Distance (EMD2)~\cite{ref-18} as the loss function for CNN training and obtained better age estimation results. Rothe et al.~\cite{ref-30} combined the VGG-16 network pretrained on ImageNet dataset, using the principal component analysis (PCA) method to obtain a lower mean absolute error (MAE) value on MORPH Album 2. Then, they transformed the age regression into an age classification problem through the Deep EXpectation (DEX) method~\cite{ref-19} and achieved better results. Recently, Hou et al.~\cite{ref-21} used the R-SAAFc2+IMDB-WIKI method to obtain best results on a FG-NET dataset. Zhang et al.~\cite{ref-22} proposed an age-and-gender estimation method combining a multi-level residual network (RoR) with two modest mechanisms, which they actively presented to achieve state-of-the-art results on the Adience benchmark. Gao et al.~\cite{ref-32} proposed a deep label distribution learning (DLDL) method, which effectively utilized the label ambiguity in both feature learning and classifier learning; thus, the best MAE value on the MORPH dataset was achieved.

\subsection{Apparent Age Estimation}
Face age estimation made a breakthroughs through the development of the convolution neural network; yet, researchers are not confined to the study of biological age estimation. Apparent age estimation was originally inspired by the 2015 ChaLearn Looking at People (LAP) competition~\cite{ref-46}, where the apparent age dataset was released. A method called logistic boosting regression (Logit Boost)~\cite{ref-33} was proposed, which realized the network optimization progressively. Xu et al.~\cite{ref-34} proposed a deep label distribution method with distribution-based loss functions and used the Coc-DPM algorithm~\cite{ref-35} and face point detector~\cite{ref-36} for face search. Zhu et al.~\cite{ref-37} used the Microsoft Project Oxford API~\cite{ref-39} and Face ++ API~\cite{ref-38} to preprocess LAP dataset, and then got GoogleNet pretrained on several other datasets. Kuang et al.~\cite{ref-40} studied the age-related discriminative performance over multiple age datasets of MORPH, FG-NET, Adience, FACES~\cite{ref-41}, and mixed with random forest and quadratic regression as well as local adjustment methods. Lin et al.~\cite{ref-42} fused real-word value-based regression models and Gaussian label distribution based classification models, which were pretrained on CASIA WebFace~\cite{ref-43}, CACD(computer aided conceptual design) ~\cite{ref-44}, WebFaceAge~\cite{ref-45} and MORPH datasets, and were fine-tuned on the ChaLearn LAP dataset. Deep EXpectation (DEX) formulation~\cite{ref-47} was proposed for apparent age estimation and won the LAP 2015 challenge. Recently, Agustsson et al.~\cite{ref-48} proposed a nonlinear regression network called Anchored Regression Network (ARN), which achieved the state-of-the-art results on 15LAP validation set.
\par
The 2016 ChaLearn LAP Apparent Age Estimation (AAE) competition~\cite{ref-56} had been completed and expanded the dataset scale based on the 15LAP dataset. Gurpinar et al.~\cite{ref-49} proposed a two-level system for estimating the apparent age of facial images, where the samples were classified into eight age groups. Duan et al.~\cite{ref-50} proposed a CNN2ELM method, where apparent age was estimated by the Race-Net + Age-Net + Gender-Net + ELM Classifier + ELM Regression (RAGN). Malli et al.~\cite{ref-51} divided the LAP dataset into age groups and age-shifted groups and used these groups to train the VGG-16 model. Uricar et al.~\cite{ref-52} extracted the deep features and formulated a SO-SVM multi-class classifier on top of it. Huo et al.~\cite{ref-53} proposed a novel method called deep age distribution learning (DADL) to use the deep CNN model to predict the age distribution. Dehghan et al.~\cite{ref-54} introduced a large dataset of 4 million face recognition images to pretrain their model, and then predicted apparent age on the age dataset. Antipov et al.~\cite{ref-55} employed different age encoding strategies for training “general” and “children” networks, including 11 ``general'' models and 3 ``children'' models, which achieved the state-of-the-art results using on 16LAP dataset.
\par
In conclusion, whether biological age estimation or apparent age estimation, one of the pivotal issues in age estimation is how to learn the distinctive features of face age.

\section{Methodology}
In this section, we describe the proposed AL-ResNets and AL-RoR architecture with the Attention LSTM network for age estimation. Our methodology is essentially composed of three steps: (1) Constructing AL-ResNets or AL-RoR architecture for improving the discrimination of the model; (2) Pretraining the CNN model of ResNets or RoR on ImageNet and fine-tuning on the IMDB-WIKI-101 dataset to alleviate over-fitting problem, and then training for global features on the target age datasets; (3) Extracting local features of the age-sensitive regions by LSTM to further improve the performance of age estimation. In the following, we will describe the three main components in detail.

\subsection{AL-ResNets and AL-RoR Architectures}
It is widely acknowledged that the performance of CNN-based age estimation relies heavily on the optimization ability of the CNN architecture, where deeper and deeper CNNs have been constructed. Particularly, ResNets~\cite{ref-14} won the first place at the ILSVRC 2015 classification task, which had achieved tremendous success in various computer vision tasks. RoR~\cite{ref-15} was constructed by adding identity shortcuts level-by-level based on original residual networks. It is noteworthy to mention that recently RoR also succeeded in the study of age estimation~\cite{ref-6}~\cite{ref-22} for its outstanding performance. In order to get state-of-the-art performance and verify the robustness of our method with different network architectures, we construct new network structures named AL-ResNets and AL-RoR based on the ResNets and RoR network architectures.
\begin{figure}
\centering
\includegraphics[width=1.0\linewidth]{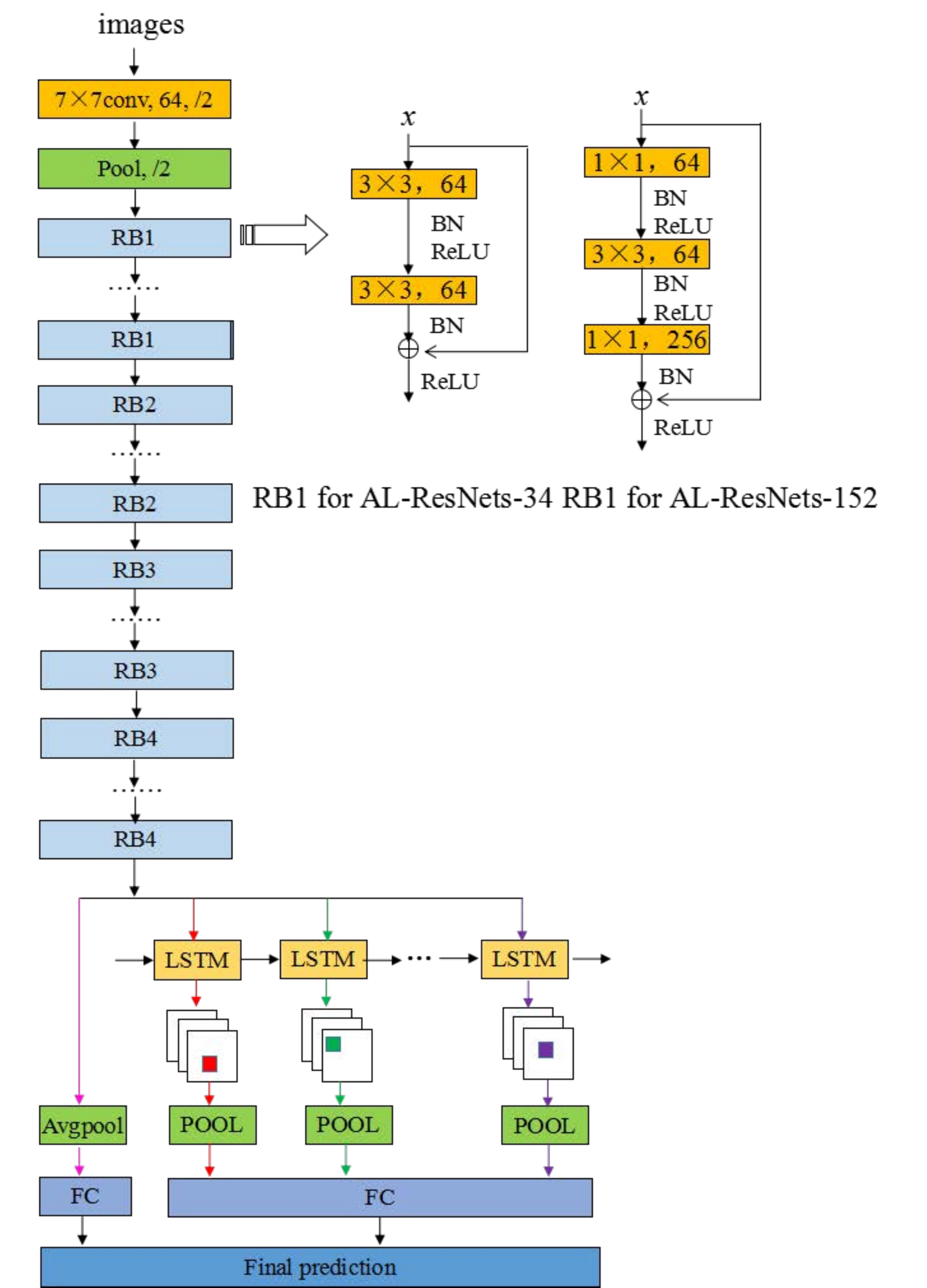}
\caption{AL-ResNets is a combination of the ResNets and LSTM unit. ResNets-34 and ResNets-152 are composed of different residual block structures. The rose branch indicates the global features extracted by ResNets on each image, and the colored areas (red, cyan, and purple) on the feature map of AL-ResNets represents the age-sensitive region extracted by LSTM on different images. Our classifier is not restricted to the raw image but rather its regional information.}
\label{fig:AL-ResNets}
\end{figure}
\par
To train the ResNets models for image classification tasks, the input RGB images need to be cast into an ordered preprocess procedure. The input images are first resized to a fixed-size of 256$\times$256, followed by a random cut to further reduce image size to 224$\times$224 before entering the network. ResNets are built on four groups of residual blocks, where their basic components (conv, BN and ReLU) operate on shortcut levels. ResNets use shortcuts to propagate information only between neighboring layers in residual blocks. The LSTM unit is not only suitable for shallow ResNets, but also fits in nicely with other various deep residual networks. As shown in Fig.~\ref{fig:AL-ResNets}, AL-ResNets-34 and AL-ResNets-152 have different residual block structures, where each residual block can be expressed in a general form:
\begin{equation}
\begin{array}{cl}
y_{l}=h(x_{l})+F(x_{l}, W_{l}),\\
x_{l+1}=f(y_{l})
\end{array}
\label{E:ResNets1}
\end{equation}
where $x_{l}$ and $x_{l+1}$ are input and output of the $l$-th block, respectively. $F$ is a residual mapping function, $h(x_{l})=x_{l}$ is an identity mapping function, and $f$ is a ReLU function.

\begin{figure}
\centering
\includegraphics[width=1.0\linewidth]{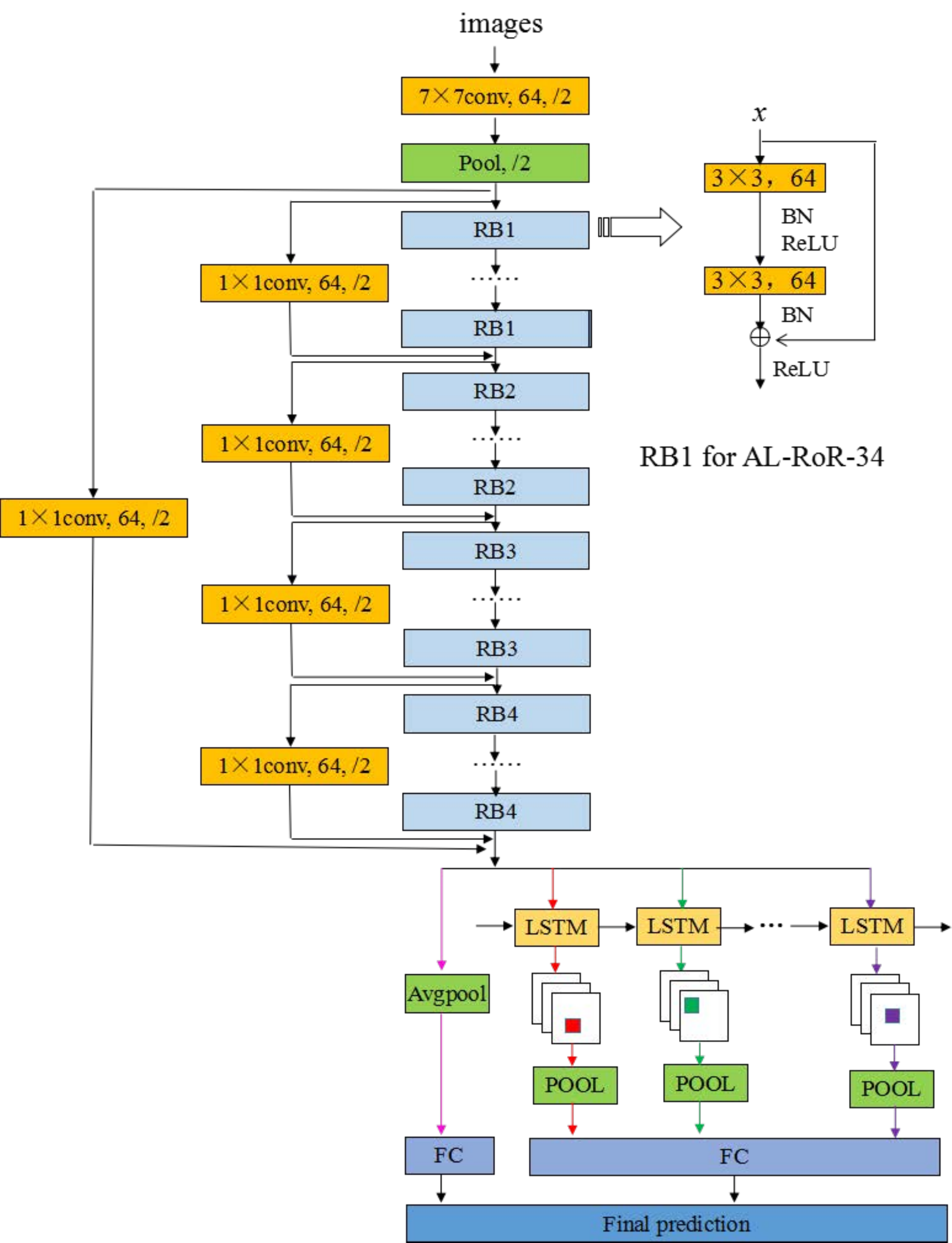}
\caption{The AL-RoR model is constructed by adding the LSTM unit to Multilevel Residual Networks. The shortcut on the left is a root-level shortcut, and the remaining shortcuts made up of the four yellow shortcuts are middle-level shortcuts. The blue shortcuts are final-level shortcuts. The LSTM unit is inserted between the residual group and the fully connected layer, which need to be applied to select the useful relative patch (red, cyan, and purple) on different images for classification.}
\label{fig:AL-RoR}
\end{figure}
ResNets transform the learning of $y_{l}$ into the learning of $F(x_{l}, W_{l})$ (single-level residual mapping) by residual block structure. Compared with ResNets, RoR~\cite{ref-15} transfers the learning problem to learning the residual mapping of residual mapping (multi-level residual mapping), which is simpler and easier than the original residual mapping to learn. In addition, RoR creates several direct paths for propagating information between different original residual blocks by adding extra shortcuts, so layers in upper blocks can propagate information to layers in lower blocks. By information propagation, RoR can alleviate the vanishing gradients problem. Fig.~\ref{fig:AL-RoR} shows the basic structure of a 34-layer AL-RoR based on RoR, which owns root-level, middle-level, and final-level shortcuts. Each residual block group contains residual blocks of 3, 4, 6 and 3 in order, and the junctions of multilevel residual mapping are located at the end of each residual block group and can be expressed by the following formulations.
\begin{equation}
\begin{array}{cl}
x_{3+1}=g(x_{1})+h(x_{3})+F(x_{3}, W_{3})\\
x_{3+4+1}=g(x_{3+1})+h(x_{3+4})+F(x_{3+4}, W_{3+4})\\
x_{3+4+6+1}=g(x_{3+4+1})+h(x_{3+4+6})\\
+F(x_{3+4+6}, W_{3+4+6})\\
x_{3+4+6+3+1}=g(x_{1})+g(x_{3+4+6+1})\\
+h(x_{3+4+6+3})+F(x_{3+4+6+3}, W_{3+4+6+3})
\end{array}
\label{E:RoR1}
\end{equation}
where $x_{l}$ and $x_{l+1}$ are input and output of the $l$-th block, and $F$ is a residual mapping function, $h(x_{l})=x_{l}$ and $g(x_{l})=x_{l}$ are both identity mapping functions. $g(x_{l})$ expresses the identity mapping of first-level and second-level shortcuts, and $h(x_{l})$ denotes the identity mapping of the final-level shortcuts.

When the images are trained on the ResNets or RoR network, effective global facial features can be obtained by extracting the output feature map of the last convolutinal layer. One of the reasonable assumptions is that the facial age prediction is not only represented by the form of global characteristics but also can be related to many age-sensitive facial parts. So it is possible to introduce the local features of age-sensitive regions to enhance discrimination for fine-grained age estimation further. In this work, we build an AL-ResNets or AL-RoR network to locate age-sensitive regions for extracting local features, which are based on the LSTM unit and CNN models (ResNets or RoR), as shown in Fig.~\ref{fig:AL-ResNets} and Fig.~\ref{fig:AL-RoR}. The output feature map of the ResNets or RoR last residual block is used as both the input of the original fully connected layer and the input of the LSTM unit. The LSTM unit locates the positions of age-sensitive regions and extracts the most significant local features for softmax classification in forward-propagation, and it is optimized in backward-propagation according to the cross entropy loss function. For combining global image-level features with local attention features in the AL-ResNets or AL-RoR network, age predictions were done by the two kinds of features separately, and then the final age prediction was determined by weighted average of the above predictions.

\subsection{Pretraining for Global Features}
In order to obtain the global and local features of facial images on Adience, MORPH Album 2, FG-NET and LAP datasets, the training phase is divided into two stages: pretraining for global features, training for local features of the age-sensitive regions. First, we extract global features by ResNets or RoR.
\par
Due to the use of small-scale target age datasets for age estimation, the over-fitting problem will occur easily if training directly on them. Drawing on the idea of transfer learning, we use ResNets or RoR network pretraining on ImageNet dataset to learn basic image feature representation, which can efficiently reduce the over-fitting problem. Moreover, the accuracy of age estimation relates to both the scale and the age distribution of the dataset. So the large-scale facial age dataset IMDB-WIKI-101 is used to fine-tune the models pretrained on ImageNet dataset for further learning the feature expression of facial age images and alleviating the over-fitting problem. IMDB-WIKI~\cite{ref-19} is the largest publicly available dataset for age estimation of people in the wild, containing more than 0.5 million images with accurate age labels. However, there are many poor-quality images in the IMDB-WIKI dataset. Zhang et al.~\cite{ref-22} first cleaned this dataset and divided them into 101 categories, and then renamed it as IMDB-WIKI-101 used to fine-tune the network models for adapting to the distribution of facial age images.
\par
In this research, the datasets for pretraining consisted of two large datasets, ImageNet and IMDB-WIKI-101. By the two stages pretraining, the model transition from general image classification to face age classification was accomplished. Finally, we fine-tuned the pretrained ResNets or RoR models on target facial age datasets (Adience, MORPH Album 2, FG-NET and LAP datasets) to get global features.

\subsection{Training For Local Features of the Age-Sensitive Regions}
The uppermost dilemma of age estimation is the similarity of the adjacent ages, we concluded that it is possible to use local features of age-sensitive regions to improve the ability of age estimation. Thus, age estimation in the wild can be treated as a fine-grained classification problem, which locates age-sensitive regions to get local features for age estimation. In this paper, we introduced the attention idea proposed by Mnih et al.~\cite{ref-27} to construct the AL-ResNets or AL-RoR network to extract local features of age-sensitive regions. Fine-grained age estimation conveniently enables part-based approaches rather than confining to global, image-level features, which improves the cohesion of contextual age information to further reduce age prediction error.
\par
\begin{figure}
\centering
\includegraphics[width=1.0\linewidth]{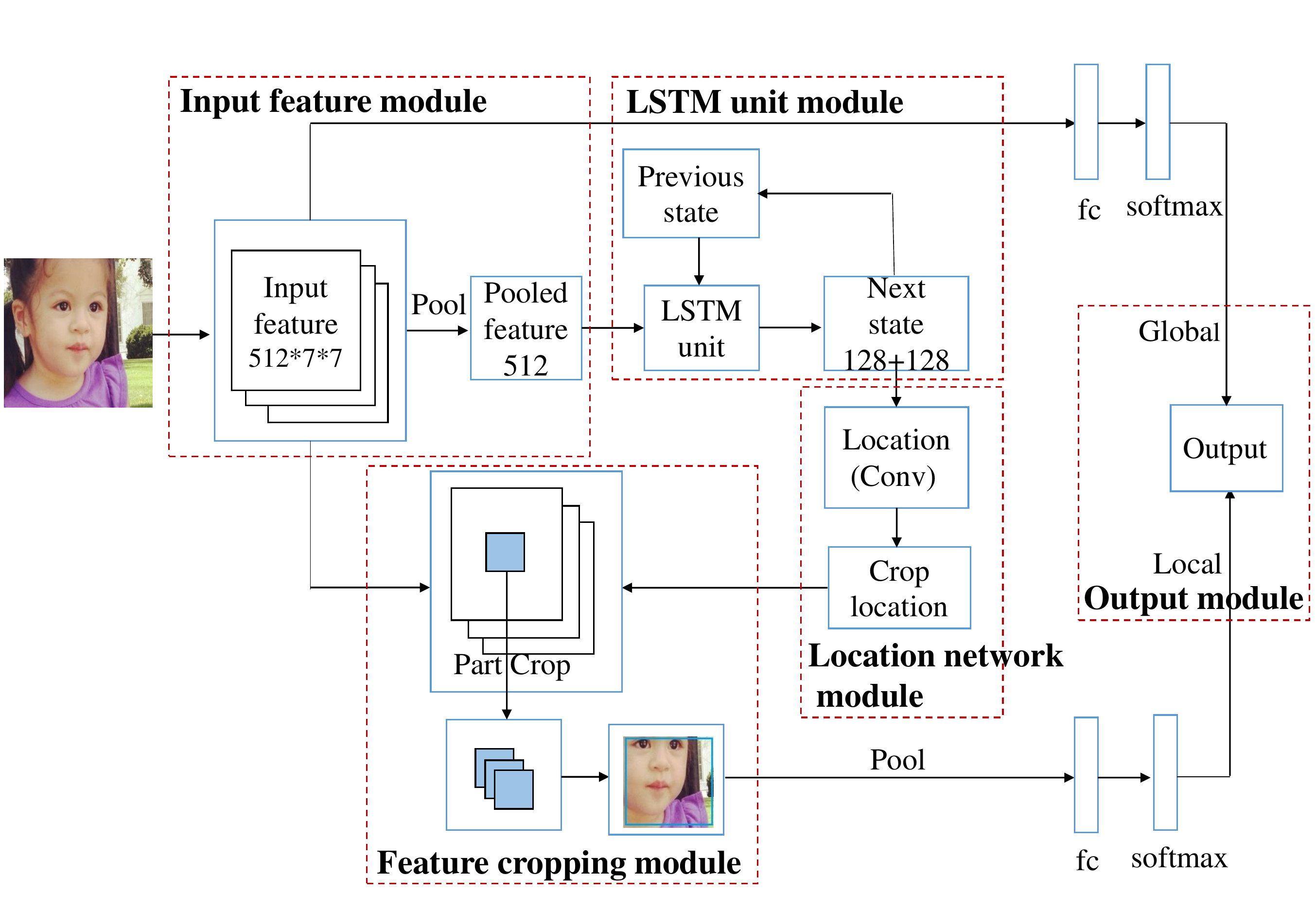}
\caption{First, the global features are extracted by input feature module; Second, LSTM unit module is used to extract the location information of age-sensitive regions; Third, the location information is used to extract the coordinate of age-sensitive region; Next, the local features are extracted by cropping the global features based on the coordinate; Finally, we calculate the final age prediction by combining the global and local age predictions in output module.}
\label{fig:local-training}
\end{figure}

The AL-ResNets or AL-RoR model is set up to automatically find age-sensitive regions and discriminative local features, and is grounded in the ResNets or RoR network to get the global features of the target age datasets. We use face global features to update the internal state of the LSTM unit and extract age-sensitive position information, and the supervised learning method is used to make our networks locate age-sensitive region automatically. In the training stage for local features of age-sensitive regions, we adopt Cross Entropy Loss as the supervised signal to train the LSTM unit module and location network module by backward-propagation algorithm. Training for local features with  the AL-ResNets or AL-RoR network consists of several parts, as shown in Fig.~\ref{fig:local-training}, which includes input feature module, LSTM unit module, location network module, feature cropping module and output module.
\par
\textbf{Input feature module:} We use the output feature map (global features) of the last residual block in the ResNets or RoR model as the input features of LSTM unit, except to reduce the possibility of local information confusion caused by over-enriching semantic information in the upper layers. There are two other reasons for using the feature map of the last convolution layer as the input of LSTM unit. The first reason is that the cropped region is a local area on the feature map, which is much smaller than the size of cropping the original image directly. So it only requires a fraction of computational cost compared with the entire network. The second reason is that cropping features on the feature map can share the same basic network so that there is no need to use a separate network training for features cropping. The size of output feature map at the last layer in the fourth residual block group of ResNets or RoR is identified as 512$\times$7$\times$7. It then goes through 7$\times$7 average pooling operation, so the LSTM unit gets a 512-dimension input based on the number of output channels. The input feature module is used to generate 512-dimensional feature vector as the input of the LSTM unit. These input features have already been trained through the basic DCNN network (ResNets or RoR), so the basic networks for global features does not employ the gradient descent algorithm at the stage of local feature training, which means that the basic networks are fixed.
\par
\textbf{LSTM unit module:} LSTM unit module is used to extract the location information of age-sensitive regions. We consider that only using the features of the current image to locate the age-sensitive region is not enough. Because the locations of age-sensitive regions in different face images are similar and regular, the location information of the age-sensitive region of other face images may be helpful to locate the age-sensitive region of the current image. Considering this assumption, we adopt LSTM unit to achieve the features for locating the age-sensitive regions. LSTM unit can automatically retain the location information of other images similar to the current image by the long-term and short-term memory mechanism. The features for locating the age-sensitive regions extracted by our Attention LSTM networks not only take into account the features of the current image, but also draw on the location information of other similar images, so these features are more comprehensive for locating the age-sensitive regions.
\par
The LSTM unit controls the cell state through the structure of ``gates,'' which is divided into input gate, forget gate, and output gate. A typical gate approach consists of two fundamental parts: a sigmoid layer and a pointwise operation. Its main implementation is as follows: First, the forget gate applies to select the information from state output at the last moment $C_{prev}$, and is followed by the input gate and a new candidate vector $C_{in-tan}$ generated by the tanh layer to create a product value. Then two sources of information are combined for status update, where the process is to abandon unnecessary information and add new information. Furthermore, the hidden layer status output of the LSTM is acquired using cell status which is maintained by the tanh layer at -1 to 1 and multiplied by the output value of the output gate. The LSTM unit performs the following computation:
\begin{equation}
\begin{array}{cl}
C_{next}=forget_{gate} \odot C_{prev}+in_{gate} \odot C_{in-tan}\\
h_{next}=out_{gate} \odot tanh(C_{next})\\
C_{in-tan}=tanh(W_{C}[h_{prev},x_{input}]+b_{C})
\end{array}
\label{E:ResNets1}
\end{equation}
where $C_{prev}$ and $C_{next}$ are the output cell state of the LSTM at the previous and current moment, respectively, $h_{next}$ is the hidden layer output state of LSTM, and all of them have the same feature dimension of 128. $C_{in-tan}$ is the candidate vector for updating the cell state. $W_{C}$ is weight parameter, and $b_{C}$ is bias. $x_{input}$ is the 512 dimensional feature as the input of the LSTM unit.
\par
The age-sensitive regions location of different faces has some similarity. By LSTM unit, both $C_{next}$ and $h_{next}$ contain long term and short-term memory information, which combine the location information of current image with previous images. LSTM unit can automatically determine the information which is more suitable for locating age-sensitive regions by the three gates. So we use both $C_{next}$ and $h_{next}$ as the location information instead of only using $x_{input}$, the 256 dimensional vector combining $C_{next}$ with $h_{next}$ is the output of LSTM unit, named as $S_{next}$.
\par
\textbf{Location network module:} The location network module consists of a convolution layer followed by a sigmoid activation function, and the output $S_{next}$ of LSTM unit is considered as the input of the convolution layer. The output of the convolution layer is a 4 dimensional vector $l_{1-4}$ as follows (4), which is used to generate the coordinate $(x,y)$, the width and height of the age-sensitive bounding box. We share the same strategy of loss functions in backward propagation process to update the LSTM unit module and location network module, as is done in the cross entropy criterion.
\begin{equation}
\begin{array}{cl}
l_{1-4}=L(W*S_{next})\\
\end{array}
\label{E:ResNets1}
\end{equation}
Where $S_{next}$ is the joint output of LSTM with a feature dimension of 256, and $W$ denotes the overall parameters. The specific form of $L(\cdot)$ is represented as a convolution layer.
\par
\textbf{Feature cropping module:} This module adopts a pooling strategy to simplify the computational complexity of the network. The module first crops the features on the  512$\times$7$\times$7 output feature map of the ResNets or RoR convolution layer according to the location coordinates. Then, an average pooling operation is performed on the cropped feature map. Finally, we get a 512 dimensional feature vector, which is the local feature of age-sensitive region.
\par
\textbf{Output module:} Finally, a key problem to solved is how to combine the global features with local features for final age prediction. In order to simply the model and reduce more training computation, we use a weighted method to combine the global prediction with local prediction in output module, and the weighted values are set to 1 and 0.5. The global features come from the feature map of the last residual block on original ResNets or RoR model, and they are used for global age prediction($P_{global}$). The local features are the features of age-sensitive regions, which are used for local age prediction($P_{local}$). The global and local features pass through their respective full connection layer and softmax layer to get global and local age prediction, respectively. The final age prediction ($P_{final}$) is the weighted average of the two prediction results, shown as:
\begin{equation}
\begin{array}{cl}
P_{final}(x)=P_{global}+0.5P_{local}\\
\end{array}
\label{E:ResNets1}
\end{equation}

\section{Experiments and Analysis}
We empirically demonstrated the effectiveness of AL-ResNets and AL-RoR on a series of benchmark datasets: Adience, MORPH, FGNET and 15/16LAP datasets. Through experiments, we analyze the effects of different training strategies, network structures and network depths to develop the best strategies, and then achieve the new state-of-the-art performance on different age datasets.

\subsection{Implementation}
The ResNets~\cite{ref-58} or RoR~\cite{ref-15} network pretrained on the ImageNet dataset is used as the fundamental model. When fine-tuning the ResNets or RoR model, the IMDB-WIKI-101 dataset is randomly divided into two parts of training and testing with the size of 90\% and 10\%, and the number of output of the softmax classifier is changed from 1000 to 101. The learning rate starts from 0.01, and is divided by a factor of 10 at epoch 60 and 90.
\par
In the target age experiments, we use the oversampling method~\cite{ref-16} by taking a ten-crop way to crop and mirror images in testing phase. We introduce deep expectation algorithm~\cite{ref-19} to deal with the problem of age regression. The network is trained for classification with M output neurons, where the number of output neurons M is set to 62, 70, and 101 for training MORPH, FGNET and LAP datasets, respectively, and where each neuron corresponds to an integer age. The weight decay is set to 1e-4 and the momentum is 0.9. The total epoch number for the Adience dataset is 60, and the learning rate starts from 0.0001. When experimenting on the MORPH Album dataset, the epoch is 120 with a learning rate of 0.001. The learning rates for the two datasets are divided by a factor of 10 after epoch 60. For training the global and local features of the FG-NET/LAP dataset, the epoch number is set to 90 and 120, respectively. The learning rate is set to 0.001, and the former is divided by a factor of 10 after epoch 30, while the latter is after epoch 60. Our implementations are based on Torch 7 with one NVIDIA GeForce GTX Titan X. For data augmentation, we all use scale and aspect ratio augmentation~\cite{ref-58}.

\subsection{Datasets}
\begin{figure}
\centering
\includegraphics[width=1.0\linewidth]{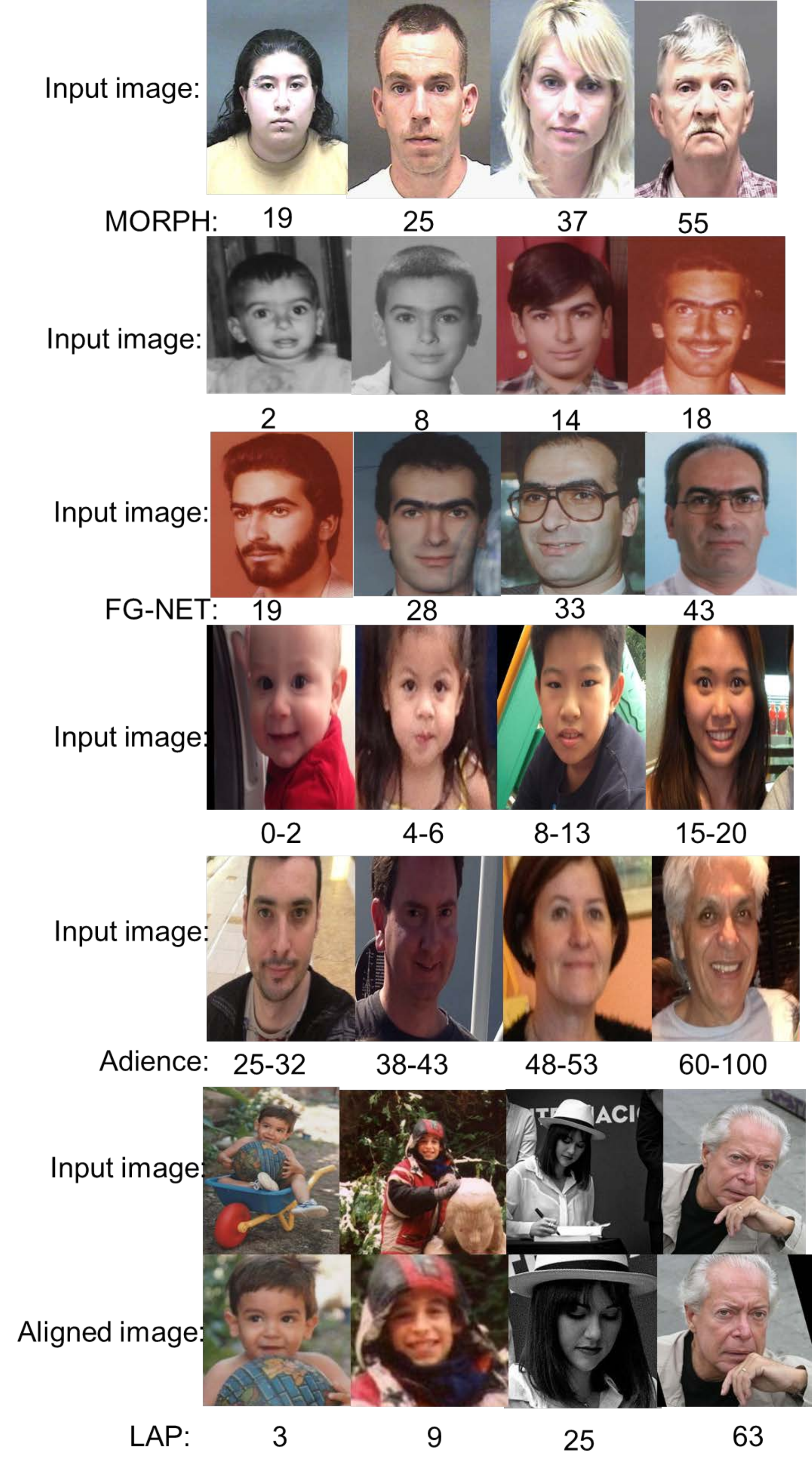}
\caption{Four different face image datasets are used in this paper. Images in MORPH and FG-NET are under constrained conditions, both of which can be applied in estimating the biological age of a person. The images in Adience and LAP datasets are in the wild, with the Adience dataset used to estimate the age group, and the LAP dataset used to predict the apparent age of a person.}
\label{fig:Example image}
\end{figure}
\textbf{MORPH:} MORPH Album 2 is one of the largest publicly available age datasets. There are 55,134 face images, whose age range from 16 to 77. This dataset contains multiple races. In order to reduce the age difference between different races or ethnic origins, we randomly selected 5,475 face images among Caucasians to avoid the influence of ethnic differences and randomly divided them into 80\% for training and 20\% for testing.
\par
\textbf{FG-NET:} FG-NET is a small dataset that only includes 1,002 images of 82 individuals in the wild, ranging from 0 to 69 years old with about 12 images per person. To ensure that everyone could provide pictures of different ages, images are collected by scanning paper documents of personal collections beside the digital images from recent years. We followed the standard leave-one-out-protocol (LOOP) for FG-NET and reported the average performance over the 82 splits. Because MORPH and FG-NET are not in the wild, we do not use any alignment method on these datasets.
\par
\textbf{Adience:} The entire Adience collection includes 26,580 256$\times$256 color facial images of 2,284 subjects, with eight age group classes (0-2, 4-6, 8-13, 15-20, 25-32, 38-43, 48-53, 60-100). Adience dataset comes from images that people automatically upload to network albums from smart phones. These images are not artificially filtered before uploading, and they are completely unconstrained as they were taken under different variations. We use the in-plane face aligned method to align faces, originally used in~\cite{ref-1}. Testing for age classification is performed using a standard five-fold, subject-exclusive cross-validation protocol, defined in~\cite{ref-16}, and the accuracy of five folds are averaged to be the final age group classification result.
\par
\textbf{LAP:} The ChaLearn 15/16LAP datasets are mainly used to study the apparent age estimation of face images. Each image label consists of the average age and the standard deviation. Most images are under unconstrained conditions (such as different background, character rotation, partial occlusion, etc.), which need face detection, alignment and cropping preprocessing. We rotated the input image in the interval of [$-60^{\circ}$, $60^{\circ}$] in $5^{\circ}$ steps and also by $-90^{\circ}$, $90^{\circ}$ and $180^{\circ}$. The face box with the strongest detection score detected by the face detector~\cite{ref-35} was taken, then the face box size was enlarged by 40\% in both width and height and the face image was cropped. For robustness, we did not delete those unaligned images, but instead, we kept them. The entire images were eventually condensed to size of 256$\times$256. The 15LAP dataset is a relatively small dataset with a total of 4,691 images, including 2,476 for training, 1,136 for validation and 1,079 for testing. The 16LAP dataset is an expanded version of 15LAP, which adds nearly 3,000 images to 15LAP.

\subsection{Evaluation Protocol}
\par
Different datasets adopt different evaluation protocols in testing phase, and there are three kinds of evaluation protocols as follows.
\subsubsection{Accuracy and 1-off Accuracy}
\par
This evaluation measures utilized in the Adience experiments are exact accuracy and 1-off accuracy, where the exact accuracy computes the correctness for the estimated age group, and the 1-off accuracy measures the results when the algorithm gives the correct or adjacent age-groups.
\subsubsection{Mean Absolute Error (MAE)}
\par
The results are evaluated in the MORPH Album 2 and FG-NET experiments by using the MAE measure. The MAE computes the error between the predicted age and the real one as follows (6), where $y_{j}$, $y_{j}^{'}$ represent the actual age and the estimated age, respectively. $N$ represents the number of all test images.
\begin{equation}
\begin{array}{cl}
MAE=\frac{1}{N}\sum_{j=1}^N\left| y_{j}-y_{j}^{'}\right|
\end{array}
\label{E:mae}
\end{equation}

\subsubsection{$\epsilon$-Error}
\par
For age estimation on LAP dataset, we employ $\epsilon$-error evaluation metric besides MAE, which is a result of the uncertainty introduced by standard deviation $\delta$. $\epsilon$-error is mainly affected by mean $\mu$ and standard deviation $\delta$, as well as the network prediction output value, where they are subject to a normal distribution. The expression of $\epsilon$-error is shown in (7), where $x_{j}$, $\mu_{j}$, $\delta_{j}$ are the predicted age, the apparent age value and the standard deviation, respectively, and $N$ is the number of all test images.
\begin{equation}
\begin{array}{cl}
\epsilon-error=\frac{1}{N}\sum_{j=1}^N(1-exp(-\frac{(x_{j}-\mu_{j})^{2}}{2\delta_{j}^{2}}))
\end{array}
\label{E:error}
\end{equation}

\subsection{Age Group Classification Experiments}
By information shortcut propagation, ResNets can alleviate the vanishing gradients problem. RoR based on ResNets benefits from the standpoint of optimization through RoR residual mapping and the extra shortcuts provided to expedite information propagation between distant layers. In order to analyze the robustness of our method in different networks, we used ResNets and RoR as the base models. To find the optimal model of the 34-layer network, we carried out a lot of comparative experiments on the Adience dataset. There are eight training methods which were used on fold4 of the Adience dataset and analyzed in terms of classification accuracy and 1-off accuracy:
\par
(1)\textbf{ ReNets-34}: Use solely Adience to train ResNets-34 network.
\par
(2) \textbf{I-ResNets-34}: After pretrained on ImageNet dataset, fine-tune the I-ResNets-34 (ImageNet-ResNets-34) network with Adience.
\par
(3) \textbf{Ft-101-RoR-34}: After pretrained on ImageNet and IMDB-WIKI-101 datasets, fine-tune the Ft-101-RoR-34 (FineTune-IMDBWIKI101-RoR-34) network with Adience.
\par
(4) \textbf{Ft-101-ResNets-34}: After pretrained on ImageNet and IMDB-WIKI-101 datasets, fine-tune the Ft-101-ResNets-34 (FineTune-IMDBWIKI101-ResNets-34) network with Adience.
\par
(5) \textbf{AL-RoR-34 (Only local features)}: Based on (3), then train the AL-RoR-34 network with Adience and predict age only based on local features.
\par
(6) \textbf{AL-ResNets-34 (Only local features)}: Based on (4), then train the AL-ResNets-34 network with Adience and predict age only based on local features.
\par
(7) \textbf{AL-RoR-34}: Based on (5), predict age based on global and local features.
\par
(8) \textbf{AL-ResNets-34}: Based on (6), predict age based on global and local features.

\begin{table}[hbp]
\renewcommand{\arraystretch}{1.3}
\caption{Age Classification Results(\%) tested on Fold4(1-crop): ResNets-34, I-ResNets-34, Ft-101-RoR-34, Ft-101-ResNets-34, AL-RoR-34(Only local features), AL-ResNets-34(Only local features), AL-RoR-34 and AL-ResNets-34 stand for eight training methods respectively.}
\label{tab:one Fold}
\centering
\begin{tabular}{|l|c|c|}
\hline
Method                        &Accuracy(\%)    &1-off(\%)   \\ \hline\hline
(1) ResNets-34                   &56.96           &90.28        \\\hline
(2) I-ResNets-34                   &60.18           &90.51         \\\hline
(3) Ft-101-RoR-34                  &65.46           &96.85          \\\hline
(4) Ft-101-ResNets-34              &65.71           &96.90          \\\hline
(5) AL-RoR-34(Only local features)                  &65.25           &96.76          \\\hline
(6) AL-ResNets-34(Only local features)              &65.33           &96.81          \\\hline
(7) \textbf{AL-RoR-34}               &\textbf{65.77}           &\textbf{97.01}           \\\hline
(8) \textbf{AL-ResNets-34}           &\textbf{66.03}           &\textbf{97.12}           \\\hline
\end{tabular}
\end{table}
\par
The results of different methods tested on fold4 of the Adience dataset are shown in Table~\ref{tab:one Fold}. We evaluated the effect of each step in the proposed method on age group classification. The learning rate of ResNets-34 begins at 0.1 and I-ResNets-34 at 0.01. For Ft-101-RoR-34, Ft-101-ResNets-34, AL-RoR-34 and AL-ResNets-34, learning rate starts from 0.0001, and all of epochs are set to 160. Compared with the result of ResNets-34, the I-ResNets-34 obtains higher accuracy because of the basic image feature expression acquisition by ImageNet pretraining. The result of Ft-101-RoR-34 or Ft-101-ResNets-34 is obviously superior to that of I-ResNets-34, which reveals that the network first pretrained by ImageNet, and then fine-tuned through the IMDB-WIKI-101 dataset~\cite{ref-22} to achieve transfer learning and alleviate the over-fitting problem, which works better than pretraining only on ImageNet.
\par
AL-ResNets-34(Only local features) and AL-RoR-34(Only local features) get good performance, which illustrate that the local features of age-sensitive regions are sensitive to predict age. The results of ResNets-34 and RoR-34 are better than AL-ResNets-34(Only local features) and AL-RoR-34(Only local features), we argue that the global features are more important than local features, so we set higher weight to the prediction with global features. AL-ResNets-34 outperforms Ft-101-ResNets-34 performance, and the same experimental performance also matches the AL-RoR-34 and Ft-101-RoR-34 model results. These results prove the robustness and effectiveness of our attention LSTM method in different networks. Since AL-ResNets-34 based on Ft-101-ResNets-34 is constructed to train the partial regions with distinctive features, improvement of age group classification results benefit from both global features and local features of the input face images. A definite improvement has come about in the ability to train effective feature vectors of the face images as a result of the proposed method. The local features of age-sensitive regions extracted by attention LSTM is efficient, which results in the further improvement on classification accuracy.
\begin{table}[!t]
\renewcommand{\arraystretch}{1.3}
\caption{Age classification results on Adience(10-crop)}
\label{tab:shallow CNN}
\centering
\begin{tabular}{|l|c|c|}
\hline
Method                        &Accuracy(\%)    &1-off(\%)   \\ \hline\hline
Ft-101-RoR-34                &66.74$\pm$2.69   &97.38$\pm$0.65  \\\hline
Ft-101-ResNets-34            &66.63$\pm$3.04   &97.20$\pm$0.65  \\\hline
\textbf{AL-RoR-34}                    &\textbf{66.82$\pm$2.79}   &\textbf{97.36$\pm$0.70}   \\\hline
\textbf{AL-ResNets-34}               &\textbf{67.47$\pm$2.83}   &\textbf{97.33$\pm$0.65}\\\hline
\end{tabular}
\end{table}
\par
From the results shown in Table~\ref{tab:one Fold}, we can see that our approach can consistently improve the age estimation performance. Therefore, further experiments on five folds of Adience use AL-ResNets-34 and AL-RoR-34 networks for age group classification, all of epochs are set to 120.  ResNets or RoR is pretrained on ImageNet first, fine-tuned on IMDB-WIKI-101 dataset and Adience dataset, then constructed using AL-ResNets-34 and AL-RoR-34 networks to train Adience. Furthermore, the over-sampling method (ten-crop) is applied on Adience dataset for better performance. Table~\ref{tab:shallow CNN} shows the effectiveness of the proposed method. From the results of Table~\ref{tab:one Fold} and Table~\ref{tab:shallow CNN}, we can see that the results of RoR are slightly worse than ResNets, which is due to the fact that the stochastic depth algorithm~\cite{ref-59} can not play a role in improving the accuracy when using large datasets to fine-tune the model; thus, RoR without a stochastic depth algorithm and ResNets had similar performances in these experiments~\cite{ref-22}.
\par
The proposed model achieves the better results with shallow AL-ResNets in terms of accuracy. We can expand it to a deeper AL-ResNets-152 network that is able to obtain more accurate age classification results under unconstrained situations; meanwhile, AL-ResNets-152 can boost to an accuracy of 67.83\% and a 1-off accuracy of 97.53\% on five folds of the Adience dataset. To sufficiently evaluate the performance of our proposed Network, we compared it with other six other state-of-the-art methods, including:
4c2f-CNN~\cite{ref-16}, R-SAAFc2~\cite{ref-17}, DEX~\cite{ref-19}, EMD~\cite{ref-18}, RSAAFc2(IMDB-WIKI)~\cite{ref-21} and RoR+IMDB-WIKI with two mechanisms~\cite{ref-22}. The age group classification results of various methods are shown in Table~\ref{tab:comparison results}. Relying on two mechanisms, gender and weight loss layers, the RoR+IMDB-WIKI~\cite{ref-22} method achieved a significant improvement in classification accuracy. However, our AL-ResNets-152 (ten-crop), trained as a single-model architecture, provided the new state-of-the-art age classification results. The success of the above experiments should be credited to the use of the AL-ResNets-152 model with attention LSTM unit, which can extract age-sensitive local facial features for age prediction. Extensive comparisons on the Adience dataset verify the effectiveness of the proposed method.
\begin{table}[!t]
\renewcommand{\arraystretch}{1.3}
\caption{The Comparison Results on Adience }
\label{tab:comparison results}
\centering
\begin{tabular}{|p{3.1cm}|c|c|}
\hline
Method                         &Accuracy(\%)      &1-off(\%)   \\ \hline\hline
4c2f-CNN~\cite{ref-16}                     &50.7$\pm$5.1         &84.7$\pm$2.2    \\\hline
R-SAAFc2~\cite{ref-17}                    &53.5                   &87.9       \\\hline
DEX w/o IMDB-WIKI
Pretrain~\cite{ref-19}                     &55.6$\pm$6.1          &89.7$\pm$1.8      \\\hline
DEX w/ IMDB-WIKI
Pretrain~\cite{ref-19}                    &64.0$\pm$4.2          &96.60$\pm$0.90     \\\hline
$VGG_{F}$-XEMD2~\cite{ref-18}                    &61.1                &94.0          \\\hline
$RES_{F}$-EMD~\cite{ref-18}                     &62.2                &94.3           \\\hline
R-SAAFc2(IMDB-WIKI)~\cite{ref-21}          &67.3                 &97.4              \\\hline
RoR34+IMDB-WIKI
with two mechanisms~\cite{ref-22}                     &66.91$\pm$2.51       &97.49$\pm$0.76           \\\hline
RoR152+IMDB-WIKI
with two mechanisms~\cite{ref-22}                      &67.34$\pm$3.56       &97.51$\pm$0.67           \\\hline
\textbf{AL-ResNets-34}           &\textbf{67.47$\pm$2.83}       &\textbf{97.33$\pm$0.65}             \\\hline
\textbf{AL-ResNets-152}           &\textbf{67.83$\pm$2.98}       &\textbf{97.53$\pm$0.59}             \\\hline
\end{tabular}
\end{table}

\subsection{Age Value Estimation Experiments}
According to the preceding experiments in above section, we analyze that our proposed model can improve the performance of age group classification task. In this section, in order to prove the generalization ability of our method, we use our proposed method with the Deep EXpectation (DEX) method~\cite{ref-19} for age regression task on other age datasets.
\par
The performance comparison on the 15LAP dataset is summarized in Table~\ref{tab:15LAP}. To evaluate the impact of the attention LSTM of our models, we trained a ResNets-34 model as a baseline using global features of the LAP dataset. The visual attention component in the AL-ResNets-34 model provides a dynamic strategy to highlight the important and discriminative region of the image. To prove its effectiveness, we compare the MAE and $\epsilon$-error results of ResNets-34 model, both of which are significantly improved. The reason is mainly derived from the fact that AL-ResNets-34 network captures the age-sensitive local features. Compared with the AL-ResNets-34 model, we can obtain superior results by AL-ResNets-152, where relative MAE and $\epsilon$-error had gains of 11.4\% and 3.57\% respectively, which shows the power of network depth and attention LSTM.
\par
The introduction of RoR can improve the optimization ability of ResNets by adding a few identity shortcuts. To achieve better age estimation results, it is important to choose a suitable RoR basic model for a satisfying performance. Because LAP datasets are relatively small, overfitting can be a critical problem for the 15LAP dataset. The over-depth of the network may also cause the overfitting problems to be even more severe on small age datasets, so we employ the shallow RoR-34 model to alleviate the overfitting problem on LAP and other small age datasets. The RoR-34 is used to train the global features and the local features of age-sensitive regions are extracted by using the attention LSTM in the AL-RoR-34 model. Our single AL-RoR-34 model achieves 0.2683 $\epsilon$-error in the validation phase, and 0.2548 $\epsilon$-error in the test phase. The MAE reaches 3.137 in the validation phase. Notably, the proposed method achieves the new state-of-the-art  results, and surpassed the winners(achieving 1st place)~\cite{ref-47} of the ChaLearn LAP 2015.
\par
\begin{figure}[h]
\centering
\includegraphics[width=1.0\linewidth]{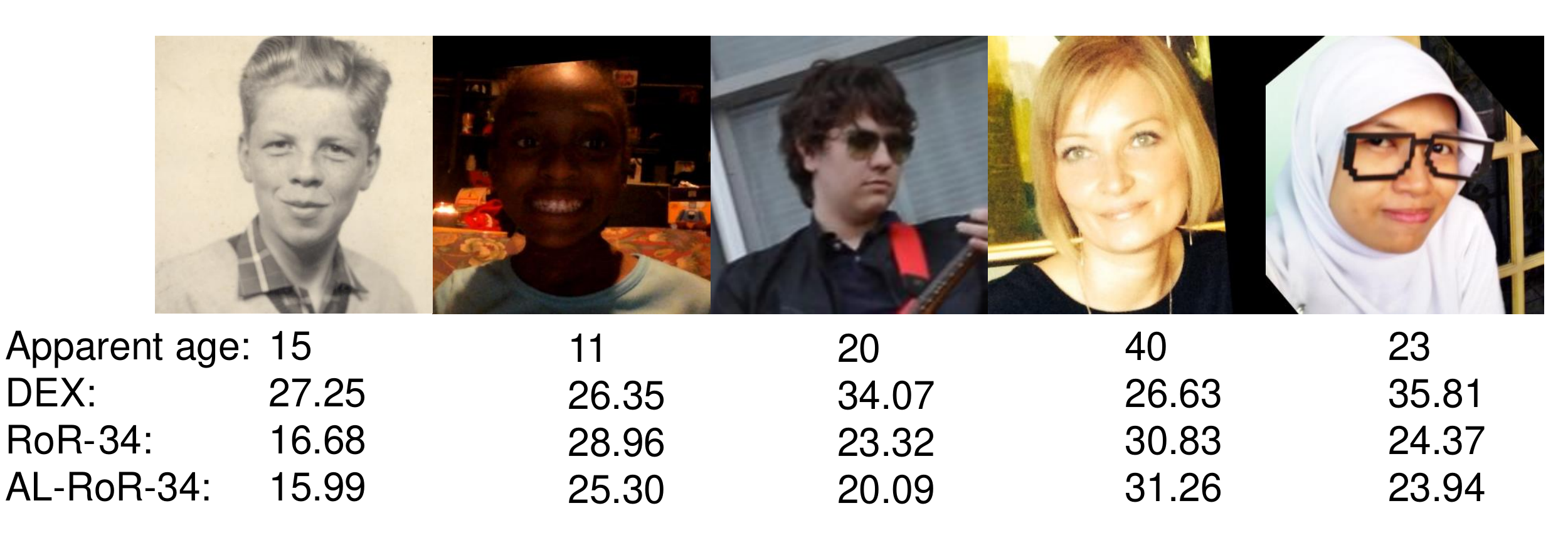}
\caption{Examples of validation images where our method significantly outperforms DEX~\cite{ref-47} and RoR-34.}
\label{fig:outperforms DEX}
\end{figure}
\begin{figure}[h]
\centering
\includegraphics[width=1.0\linewidth]{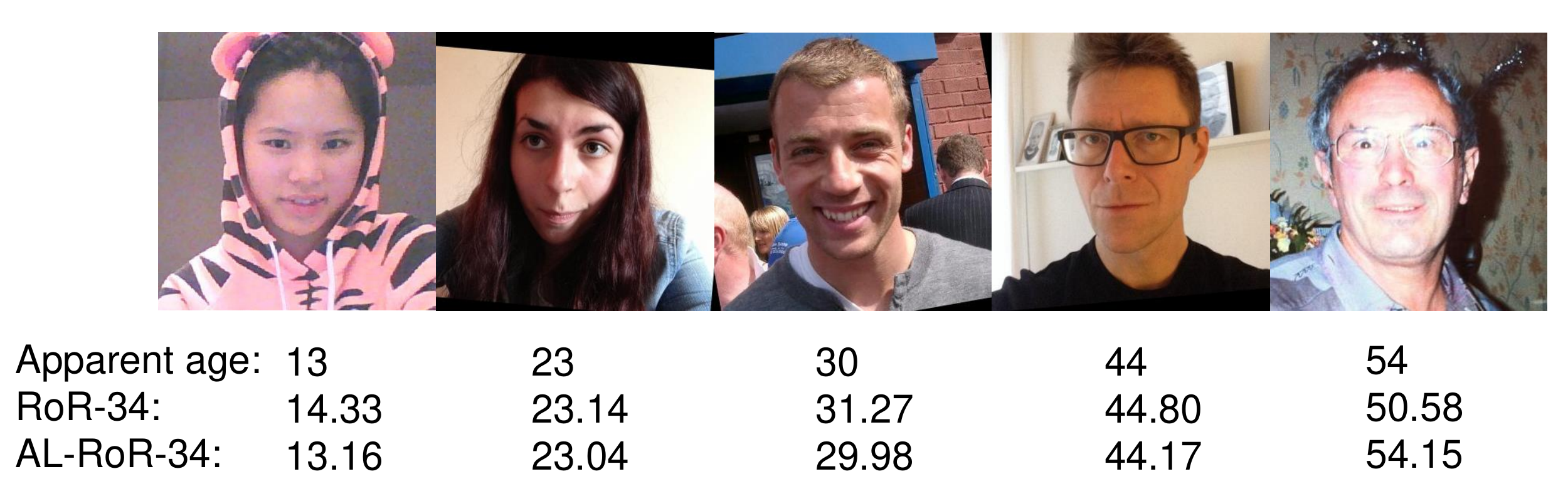}
\caption{Examples of validation images where our proposed method obtained the smallest absolute error.}
\label{fig:smallest absolute error}
\end{figure}

Fig.~\ref{fig:outperforms DEX} shows some representative examples of validation images on 15LAP dataset where our method significantly outperforms DEX~\cite{ref-47}, and the error ranges of age predictions can be greatly reduced compared with RoR-34. Some validation images with the minimum age prediction errors are shown in Fig.~\ref{fig:smallest absolute error}, which indicates that the more accurate age values than RoR-34 can be obtained in different age ranges by our proposed method. From Fig.~\ref{fig:outperforms DEX} and Fig~\ref{fig:smallest absolute error}, we find that the age predictions by AL-RoR-34 are more accurate than RoR-34, because the local features of age-sensitive regions play an important role for age estimation.
\begin{table}[!t]
\renewcommand{\arraystretch}{1.3}
\caption{The comparison results on 15LAP}
\label{tab:15LAP}
\centering
\begin{tabular}{|p{2.7cm}|p{1.4cm}|p{1.4cm}|p{1.6cm}|}
\hline
Method                            &MAE(Val)       &$\epsilon$-error(Val)      &$\epsilon$-error(Test)           \\ \hline\hline
Logit Boost~\cite{ref-33}                   &7.2949       &0.5483             &--         \\\hline
DADL~\cite{ref-34}                          &--           &0.3806             &0.3057      \\\hline
age group~\cite{ref-37}                     &--           &0.3162             &0.2948       \\\hline
Rich Coding~\cite{ref-40}                  &3.29         &0.3273              &0.2872        \\\hline
AgeNet~\cite{ref-42}                        &3.334        &0.2922             &0.2706         \\\hline
DEX1~\cite{ref-19}                          &3.252        &0.282              &0.2649          \\\hline
DEX2~\cite{ref-47}                          &3.221        &0.278              &0.2649           \\\hline
ARN~\cite{ref-48}                           &3.153        &--                 &--                \\\hline
ResNets-34                        &3.712        &0.3167             &0.3003             \\\hline
\textbf{AL-ResNets-34}                     &\textbf{3.357}        &\textbf{0.2810}             &\textbf{0.2772}              \\\hline
\textbf{AL-ResNets-152}                    &\textbf{3.243}       &\textbf{0.2778}             &\textbf{0.2668}               \\\hline
\textbf{AL-RoR-34}                         &\textbf{3.137}        &\textbf{0.2683}             &\textbf{0.2548}                \\\hline
\end{tabular}
\end{table}
\par
We also performed an additional experiment on the 16LAP dataset to demonstrate the superiority of our model. Table~\ref{tab:16LAP} summarizes the results for the 2016 ChaLearn challenge on apparent age estimation. Our AL-RoR-34 model achieved a test $\epsilon$-error of 0.2859 thereby obtaining the second place. Our result is slightly worse than OrangeLabs~\cite{ref-55}. This is because we rounded the label value on the 16LAP dataset to satisfy classification requirements, but it has an impact on the range of predicted age errors. In addition, OrangeLabs~\cite{ref-55} introduced an additional private dataset to address the shortcomings in the supply of children images, and the final test result depended on the combined results of multiple models, while our results are based only on one 34-layer model.
\begin{table}[!t]
\renewcommand{\arraystretch}{1.3}
\caption{The comparison results on 16LAP }
\label{tab:16LAP}
\centering
\begin{tabular}{|p{3.0cm}|p{1.8cm}|p{1.8cm}|}
\hline
Method                           &$\epsilon$-score(Test)   &single model?  \\ \hline\hline
DeepAge                         &0.4573                   &Yes                \\\hline
MIPAL SNU                       &0.4569                   &No                 \\\hline
Bogazici~\cite{ref-49}          &0.3740                   &No                \\\hline
CNN2ELM~\cite{ref-50}           &0.3679                   &No                \\\hline
ITU SiMiT~\cite{ref-51}         &0.3668                   &No                \\\hline
WYU CVL                         &0.3405                   &No                \\\hline
cmp+ETH~\cite{ref-52}           &0.3361                   &No               \\\hline
palm seu~\cite{ref-53}          &0.3214                   &No              \\\hline
DNN~\cite{ref-54}               &0.319                    &Yes             \\\hline
OrangeLabs~\cite{ref-55}        &0.2411                   &No            \\\hline
\textbf{AL-RoR-34}                       &\textbf{0.2859}                   &\textbf{Yes}            \\\hline
\end{tabular}
\end{table}

\begin{table}[!t]
\renewcommand{\arraystretch}{1.3}
\caption{Comparison results (MAE) for age estimation on MORPH Album 2 and FG-NET}
\label{tab:morph-fgnet}
\centering
\begin{tabular}{|l|c|c|}
\hline
Method                                        &MORPH Album 2    &FG-NET  \\ \hline\hline
AGES~\cite{ref-23}                            &8.83             &6.77    \\\hline
OHRank~\cite{ref-24}                          &6.07             &4.48 \\\hline
CA-SVR~\cite{ref-26}                          &5.88             &4.67 \\\hline
DLA~\cite{ref-25}                             &4.77             &4.26 \\\hline
BIF+KCCA~\cite{ref-28}                        &3.98             &-- \\\hline
CNN (Multi-Task)~\cite{ref-29}                &3.63             &-- \\\hline
MF+VisReg~\cite{ref-30}                       &3.45             &-- \\\hline
DEX~\cite{ref-19}                             &3.25             &4.63 \\\hline
DEX(IMDB-WIKI)~\cite{ref-19}                  &2.68             &3.09 \\\hline
R-SAAFc2(IMDB-WIKI)~\cite{ref-21}             &--               &3.01 \\\hline
DLDL~\cite{ref-32}                            &2.42             &-- \\\hline
\textbf{AL-RoR-34}                                     &\textbf{2.36}             &\textbf{2.39} \\\hline
\end{tabular}
\end{table}
The proposed model also gave a superior performance on MORPH Album 2 and FG-NET datasets, where the attention LSTM network is essential for age estimation too. As shown in Table~\ref{tab:morph-fgnet}, by training on MORPH Album 2 dataset using the AL-RoR-34 network, the best MAE value of 2.36 years is achieved, which is an improvement of 0.06 over the DLDL~\cite{ref-32} methods. On FG-NET dataset, we achieve a new state-of-the-art MAE of 2.39 years. Compared to only using global features on both datasets, the addition of local features further reduces the age prediction error. All these experiments on different age datasets show that combining global and local features can result in better performance for age estimation.

\section{Conclusion}
This paper proposes an AL-ResNets and an AL-RoR architectures based on the attention LSTM network for the task of facial age estimation. Fine-grained age estimation method effectively learns the discriminative local features of the age-sensitive regions obtained by the attention LSTM unit. It combines the global and the local features on the target age datasets to achieve better performance. Pretraining on ImageNet is used to learn the basic image feature representation. Further fine-tuning on IMDB-WIKI-101 helps to learn the feature expression of the facial age images. By introducing the fine-grained classification and the visual attention mechanism into the age estimation task, we not only obtain the state-of-the-art performance on Adience, MORPH, FGNET and 15/16LAP datasets, but also provide new feasible ideas for face age estimation and face analysis research.

% or


\appendices

\ifCLASSOPTIONcaptionsoff
  \newpage
\fi

\end{document}